\documentclass[10pt,twocolumn,letterpaper]{article}

\usepackage{multirow}  
\usepackage{geometry}  
\usepackage{array}     
\usepackage{graphicx}  

\usepackage{amsmath}
\usepackage{amssymb}
\usepackage{mathtools}
\usepackage{amsthm}
\usepackage{pifont}    
\usepackage{algorithm}
\usepackage{algpseudocode}

\usepackage{iccv}              
\definecolor{iccvblue}{rgb}{0.21,0.49,0.74}
\usepackage[pagebackref,breaklinks,colorlinks,allcolors=iccvblue]{hyperref}

\def\paperID{11467} 
\def\confName{ICCV}
\def\confYear{2025}

\title{Outdoor Monocular SLAM with Global Scale-Consistent 3D Gaussian Pointmaps}

\author{Chong Cheng\textsuperscript{1*}
\and Sicheng Yu\textsuperscript{1*}
\and Zijian Wang\textsuperscript{1} 
\and Yifan Zhou\textsuperscript{1} 
\and Hao Wang\textsuperscript{1\dag}
\and \textsuperscript{1}The Hong Kong University of Science and Technology (Guangzhou)
\and {\small \texttt{ccheng735@connect.hkust-gz.edu.cn}} \quad
{\small \texttt{yusch@mail2.sysu.edu.cn}} \quad
\and {\small \texttt{zwang886@connect.hkust-gz.edu.cn}} \quad
{\small \texttt{yzhou223@jhu.edu}} \quad
{\small \texttt{haowang@hkust-gz.edu.cn}}
}
\vspace{-20pt}
\begin{document}
\twocolumn[{
\maketitle
\begin{center}
    \vspace{-15pt}
    \includegraphics[width=\textwidth]{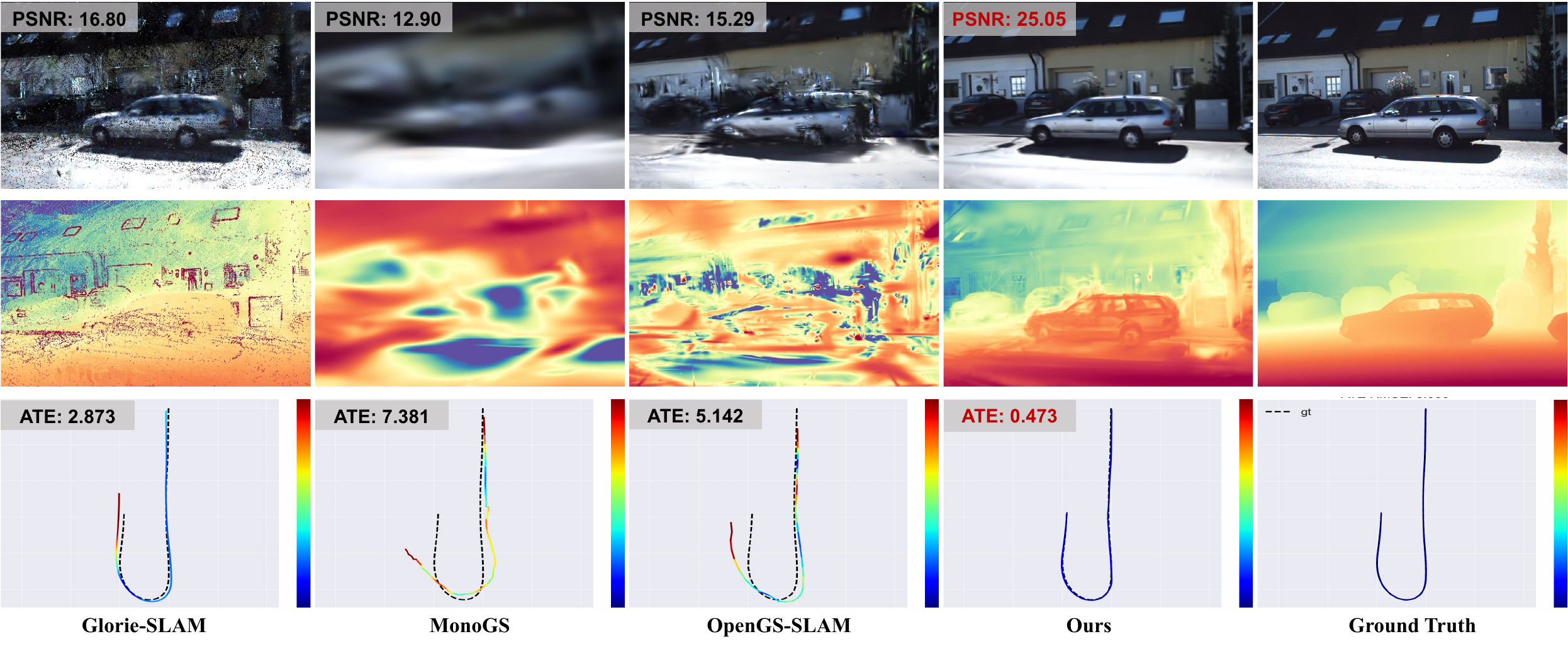}
    \label{fig:teaser}
    \vspace{-20pt}
    \captionof{figure}{
    \textbf{Localization and novel view synthesis results on KITTI.} Our method S3PO-GS maintains robust tracking and high-quality novel view synthesis even in cases of large-angle turns. This is achieved through our self-consistent 3DGS pointmap tracking and the patch-based pointmap dynamic mapping module.
}
\end{center}
}]
\renewcommand{\thefootnote}{}
\footnotetext{* Equal contribution.}
\footnotetext{\dag~Corresponding author.}

\begin{abstract}
3D Gaussian Splatting (3DGS) has become a popular solution in SLAM due to its high-fidelity and real-time novel view synthesis performance.
However, some previous 3DGS SLAM methods employ a differentiable rendering pipeline for tracking, \textbf{lack geometric priors} in outdoor scenes. Other approaches introduce separate tracking modules, but they accumulate errors with significant camera movement, leading to \textbf{scale drift}. To address these challenges, we propose a robust RGB-only outdoor 3DGS SLAM method: S3PO-GS.
Technically, we establish a self-consistent tracking module anchored in the 3DGS pointmap, which avoids cumulative scale drift and achieves more precise and robust tracking with fewer iterations.
Additionally, we design a patch-based pointmap dynamic mapping module, which introduces geometric priors while avoiding scale ambiguity. This significantly enhances tracking accuracy and the quality of scene reconstruction, making it particularly suitable for complex outdoor environments.
Our experiments on the Waymo, KITTI, and DL3DV datasets demonstrate that S3PO-GS achieves state-of-the-art results in novel view synthesis and outperforms other 3DGS SLAM methods in tracking accuracy. Project page: \url{https://3dagentworld.github.io/S3PO-GS/}.
\end{abstract}    
\section{Introduction}
\label{sec:intro}
Visual Simultaneous Localization and Mapping (SLAM), a core problem in fields like autonomous driving, robotics, and virtual reality (VR), has received substantial attention. Within this area, 3D scene representation has become a primary research focus, resulting in the development of numerous sparse \cite{mur2015orb, Campos_2021, mur2017orb, lategahn2011visual} and dense \cite{newcombe2011kinectfusion, newcombe2011dtam, zhang2024glorie} representation methods, which advance localization accuracy. However, these methods still face significant challenges in novel view synthesis (NVS) capabilities.

Given the photorealistic visual effects offered by 3D Gaussian Splatting (3DGS) \cite{kerbl20233dgs} scene representations, recent research has focused on integrating 3DGS with SLAM \cite{matsuki2024gaussian, huang2024photo, yan2024gs,keetha2024splatam, hu2024cg-slam, yu2025opengs-slam}. However, existing 3DGS SLAM methods still face two key challenges in outdoor RGB-only scenarios: \textbf{lack of geometric priors} and \textbf{scale drift} issues.

On one hand, some previous RGB-only 3DGS SLAM methods, such as those proposed by \cite{matsuki2024gaussian}, perform pose estimation via differentiable rendering pipelines. However, this approach \textbf{lacks geometric priors} and struggles with convergence in complex environments, particularly in outdoor settings, where the model is prone to getting stuck in local minima.

On the other hand, to enforce geometric constraints, some methods \cite{huang2024photo, yu2025opengs-slam,zhu2024mgs-slam} introduce independent tracking modules and pre-trained models to supplement geometric information, enhancing the robustness of pose estimation. Yet, this strategy requires maintaining scale alignment between external modules and the 3DGS map. In scenarios with large rotations and displacements, accumulated errors can easily lead to \textbf{scale drift} in SLAM system, degrading subsequent pose estimation and map reconstruction quality.

To address the challenges above, we propose a robust 3D Gaussian Splatting SLAM method---\textbf{S3PO-GS}. 
Our approach leverages pre-trained pointmap models to compensate for the lack of geometric priors in RGB-only scenarios. By anchoring 3DGS-rendered pointmaps, we establish 2D-3D correspondences, enabling scale self-consistent pose estimation. Through a patch-based design, we align the scale of the pre-trained pointmap with the current 3DGS scene. This allows us to incorporate geometric priors while effectively avoiding the issue of scale drift.

Technically, we first design a self-consistent 3DGS pointmap tracking module that estimates poses through pixel-wise 2D-3D correspondences between the input frame and 3DGS-rendered pointmap.  
The pre-trained model serves solely as a bridge for correspondence without participating in the pose estimation, inherently avoiding scale alignment issues.
Combined with the 3DGS differentiable pipeline to optimize poses, even in complex outdoor environments, this approach can achieve more accurate and robust tracking with only 10\% of the iterations required.

Furthermore, to address the lack of geometric priors in monocular SLAM, we design a patch-based pointmap dynamic mapping. This approach employs a patch-scale alignment algorithm to achieve local geometric calibration between the pre-trained pointmap and the 3DGS scene. A dynamic pointmap replacement mechanism is designed to reduce reconstruction errors. These strategies introduce geometric priors and resolve the issue of scale ambiguity, enabling high-quality scene mapping.

Experiments on Waymo \cite{Sun_2020_CVPRwaymo}, KITTI \cite{geiger2013kittidata}, and DL3DV \cite{ling2024dl3dv} datasets show that S3PO-GS outperforms existing 3DGS SLAM methods. It also sets new benchmarks in tracking accuracy and novel view synthesis. Our main contributions include:
\begin{itemize}

\item We propose a self-consistent 3DGS pointmap tracking module that introduces priors while avoiding scale alignment issues, enhancing tracking accuracy and robustness with a significant reduction in iterations.

\item Our proposed patch-based pointmap dynamic mapping module leverages a pre-trained model to dynamically adjust the 3DGS pointmap while mitigating scale ambiguities, significantly improving scene reconstruction quality.

\item Evaluations on multiple datasets demonstrate that our method establishes state-of-the-art performance in tracking accuracy and novel view synthesis within the 3DGS SLAM framework.

\end{itemize}

\section{Related work}

\subsection{Classical SLAM}
Classical SLAM methods commonly use sparse feature representations. For example, methods in the ORB-SLAM series \cite{mur2015orb, Campos_2021,mur2017orb} combine the FAST corner detector and BRIEF descriptor, tracking and updating only a small number of key points. Similarly, SIFT \cite{lowe2004distinctive} and SURF \cite{bay2006surf} also rely on feature points for camera pose estimation. Based on this efficient feature tracking idea, PTAM \cite{klein2007parallel} first parallelizes tracking and mapping, marking the beginning of real-time keypoint-based SLAM research. However, the resulting maps are typically sparse, serving primarily for navigation and localization rather than detailed scene modeling.

Dense SLAM \cite{newcombe2011kinectfusion, newcombe2011dtam, zhang2024glorie} generates detailed 3D maps, contrasting with sparse methods focused on pose estimation, and is well-suited for augmented reality and robotics. It includes frame-centered approaches, which are efficient but struggle with global consistency, and map-centered approaches using voxel grids or point clouds to enhance tracking and system compactness \cite{prisacariu2014framework, inproceedings}. Recent advancements like iMAP \cite{sucar2021imap} integrate neural networks for enhanced detail capture but face significant computational demands, limiting real-time applications. 
GlORIE-SLAM \cite{zhang2024glorie} utilizes a flexible neural point cloud representation, improving real-time performance without the need for costly backpropagation. However, it still does not achieve photorealistic novel view synthesis.

\subsection{NeRF-based and 3DGS-based SLAM}
NeRF \cite{mildenhall2020nerfrepresentingscenesneural} uses a Multi-Layer Perceptron (MLP) to sample along viewing rays and generate high-quality novel view synthesis via volume rendering, significantly outperforming traditional sparse SLAM methods in reconstruction accuracy \cite{sandström2023pointslamdenseneuralpoint,wang2023coslam,johari2023eslam,yang2022voxfusion}. In the SLAM framework, NeRF optimizes the MLP using multi-view geometric information to achieve high-fidelity scene representation \cite{rosinol2022nerfslamrealtimedensemonocular,zhu2022niceslam,zhu2024nicer}. However, its long training time limits applicability in real-time SLAM \cite{garbin2021fastnerf}. Recent research introduce explicit structures like multi-resolution voxel grids or hash encodings to improve rendering speed and efficiency \cite{muller2022instant,hu2022efficientnerf}, yet they still struggle with achieving real-time rendering. 

\begin{figure*}
  \centering
  \includegraphics[width=0.95\textwidth]{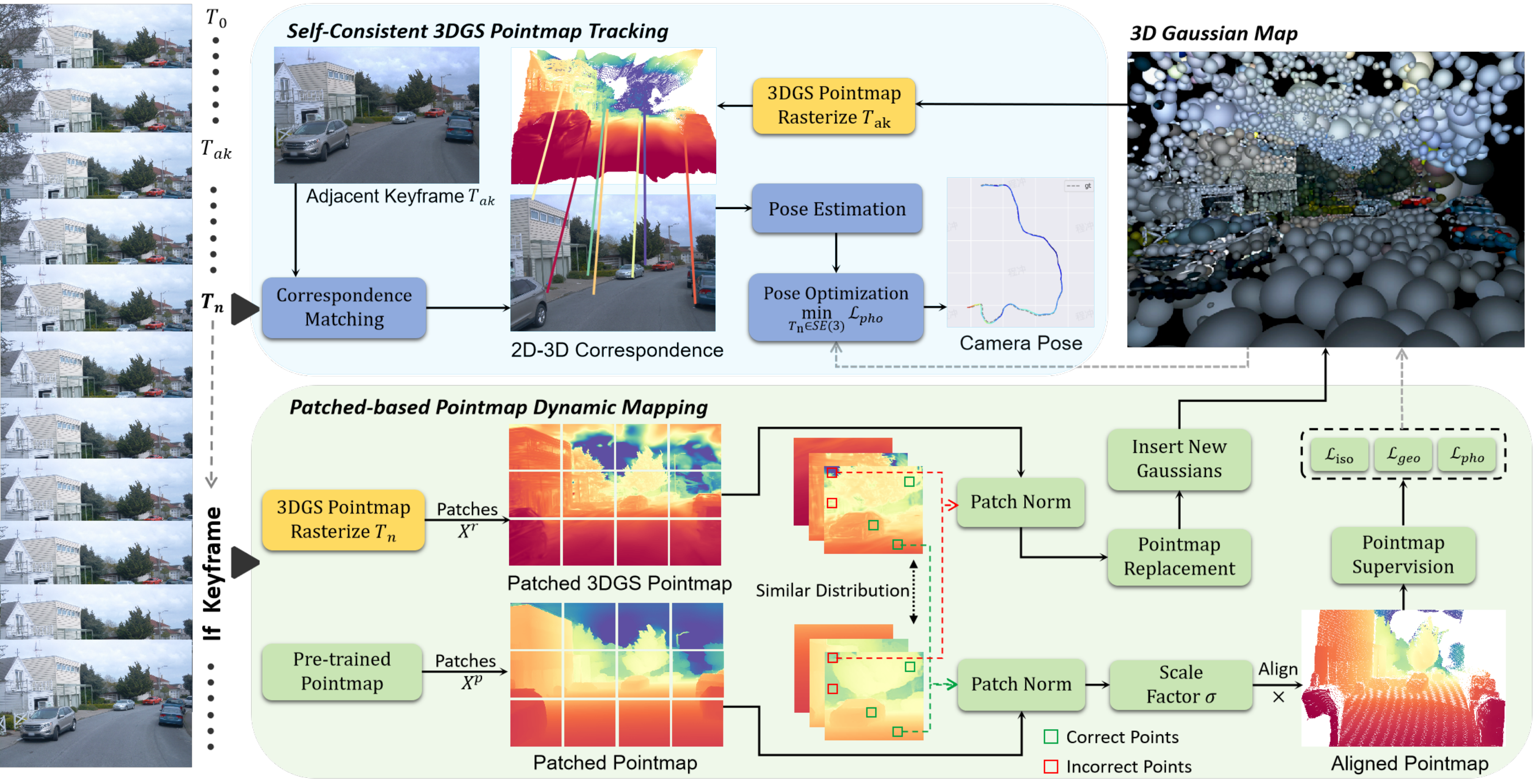}
  \vspace{-7pt}
  \caption{\textbf{S3PO-GS pipeline for SLAM.} 
  The system begins by initializing a 3D Gaussian map (optimizing MASt3R's pointmap for 1000 steps). For new input frame \(T_n\), we rasterize the 3DGS pointmap of the adjacent keyframe \(T_{ak}\), match it with the input image, and establish 2D-3D correspondences to estimate scale self-consistent pose. The estimated pose is further refined using photometric loss. If \(T_n\) is selected as keyframe, we obtain its rendered pointmap \(X^r\) and pre-trained pointmap \(X^p\), then crop both into patches with similar distributions. After patch normalization, the correct points are selected to compute a scaling factor, which is then used to adjust \(X^p\). Once the incorrect points are replaced, \(X^r\) is used to insert new Gaussians. Finally, the aligned pre-trained pointmap is used to jointly optimize the 3D Gaussian map, enabling precise and robust localization and mapping.
  }
  \label{fig:framework}
  \vspace{-10pt}
\end{figure*}

3DGS-based  \cite{kerbl20233dgs} SLAM methods \cite{matsuki2024gaussian, yan2024gs,keetha2024splatam, hu2024cg-slam, yu2025opengs-slam} employ explicit 3D Gaussian representations for modeling and rendering scenes. Compared to traditional point cloud representations, they enable real-time scene reconstruction and provide high-fidelity view synthesis \cite{chen2024survey}. 
However, this method lacks geometric priors and struggles in outdoor environments with only RGB, requiring numerous iterations and often failing to converge. Additionally, some 3DGS-SLAM methods \cite{huang2024photo, zhu2024mgs-slam,yu2025opengs-slam} decouple camera tracking from scene modeling, using an independent model for pose estimation while relying on a 3D Gaussian distribution for reconstruction. However, these methods require maintaining a scale factor, which easily accumulates scale errors in outdoor scenes with large angular movements. This leads to scale drift, degrading both localization accuracy and reconstruction quality.

We propose a 3DGS-based approach that enables efficient, accurate, and robust tracking with RGB-only input, while achieving high-fidelity novel view synthesis.

\section{Method}
Our method comprises three main parts: 3D Gaussian Splatting (3DGS), Self-Consistent 3DGS Pointmap Tracking and Patch-based Pointmap Dynamic Mapping, as illustrated in \cref{fig:framework}. These components together form our 3DGS SLAM pipeline.

\subsection{3D Gaussian Splatting}
\label{3D Gaussian Splatting}
We employ a 3DGS \cite{kerbl20233dgs} scene representation, where the scene is modeled using a set of Gaussians centered at points $\mu$, each defined by its covariance matrix $\Sigma$:
\begin{equation}
\Sigma = RSS^T R^T,
\end{equation}
where $R$ is a rotation matrix and $S$ is a scaling matrix. By projecting these 3D Gaussians onto a 2D plane and applying tile-based rasterization, we achieve efficient and differentiable rendering on a CUDA pipeline. Unlike the original 3DGS method, we do not employ spherical harmonics; instead, we directly compute the color of pixel $x'$ using:
\begin{equation}
C(x') = \sum_{i \in N} c_i \alpha_i \prod_{j=1}^{i-1} (1 - \alpha_j),
\end{equation}
where $N$ is the set of Gaussians affecting $x'$, and $\alpha_i$ is opacity. This method allows our SLAM system to optimize all Gaussian parameters, including position, rotation, scale, opacity, and color.

\subsection{Self-Consistent 3DGS Pointmap Tracking}

\subsubsection{Pointmap Anchored Pose Estimation}
\label{Pointmap Anchored Pose Estimation}
Previous 3DGS SLAM methods \cite{yu2025opengs-slam}, relying on pre-trained modules for direct pose estimation, often encounter cumulative scale drift despite scale alignment techniques. To address this challenge, particularly in outdoor scenes with large rotation angles and displacements, we propose a novel pose tracking method. Inspired by visual localization \cite{wang2024dust3r, revaud2023sacregsceneagnosticcoordinateregression, brachmann2018dsacdifferentiableransac, Yang2019SANetSA}, we introduce a differentiable pointmap rendering pipeline within the 3DGS framework. This pipeline captures normalized 3D shape and viewpoint information through 3DGS-rendered pointmaps, establishing a basis for a scale self-consistent tracking module.

Our core innovation lies in estimating poses directly from the 3DGS scene's own scale, through the pixel-to-point 2D-3D correspondence between the 3DGS-rendered pointmaps and new input frame. Notably, the pre-trained pointmap model \cite{wang2024dust3r, mast3r} is used solely to establish these correspondences and does not directly contribute to the estimation process. 

Specifically, starting from the adjacent keyframe \(I_{ak}\), we leverage the 3DGS rasterization mechanism to build a differentiable pointmap rendering pipeline. Using the adjacent keyframe's viewpoint \(\mathbf{T}_{ak} \in SE(3)\), we render a depth map \(D_{ak} \in \mathbb{R}^{W \times H}\). The depth value at pixel \((i,j)\) is computed by alpha-blending Gaussian primitives along the ray:
\begin{equation}
D_{ak}(i,j) = \sum_{k \in N} z_k \alpha_k \prod_{l=1}^{k-1} (1 - \alpha_l),
\end{equation}
where \(z_k\) is the distance from point \(u_i\) to the camera center, \(N\) denotes the set of Gaussian primitives along the ray sorted by $z_k$, and \(\alpha_k\) represents the opacity. The rendered pointmap \(X^r_{ak}\) is derived from the depth map $D_{ak}$, by applying the inverse of the camera intrinsic matrix $K$ to each pixel's depth-scaled coordinates:
\begin{equation}
\resizebox{\columnwidth}{!}{
  $X^r(i,j) = K^{-1} \begin{bmatrix} i D(i,j) \\ j D(i,j) \\ D(i,j) \end{bmatrix} = \begin{bmatrix} \frac{1}{f_x} & 0 & -\frac{c_x}{f_x} \\ 0 & \frac{1}{f_y} & -\frac{c_y}{f_y} \\ 0 & 0 & 1 \end{bmatrix} \begin{bmatrix} i D(i,j) \\ j D(i,j) \\ D(i,j) \end{bmatrix}$},
\end{equation}
simplifying this multiplication yields:
\begin{equation}
X^r(i,j) = \begin{bmatrix} \frac{i D(i,j) - c_x D(i,j)}{f_x} \\ \frac{j D(i,j) - c_y D(i,j)}{f_y} \\ D(i,j) \end{bmatrix},
\end{equation}
where \(f_x\),\(f_y\) are the focal lengths, and \(c_x\),\(c_y\) are the optical centers of camera. From this, we construct the rendered pointmap \(X_{ak}^r\), which store viewpoint and shape information. Using $D_{ak}$ as a bridge, we establish point-to-pixel correspondences between the pointmap and image: 
\begin{equation}\label{eq:corr1}
    X_{ak}^r \leftrightarrow D_{ak} \leftrightarrow I_{ak},
\end{equation}
where the scale of \(X_{ak}^r\) originates from the 3DGS map, independent of external dependencies.

Meanwhile, we input the current frame image $I_{n}$ and the adjacent keyframe image $I_{ak}$ into the pre-trained model \cite{mast3r,wang2024dust3r} to obtain two sets of pointmaps $X^{p}_{ak},X^p_{n}$ at the same scale, along with confidence scores $c$. Based on this, we establish pointmap correspondences \(X_{ak}^p \leftrightarrow X_{n}^p\) by minimizing the pointmap distance and filter out low-confidence points \cite{mast3r}:
\begin{equation}
\resizebox{\columnwidth}{!}{
  $(i,j) \leftrightarrow (u,v) \,\big|\, \underset{(i,j),(u,v)}{\arg\min}\, \sum_{\substack{c(i,j)\geq t\\ c(u,v)\geq t}}\| X^p_{ak}(i,j) - X^p_{n}(u,v) \|$},
\end{equation}
where \(X^p(u,v) \in \mathbb{R}^3\) denotes the 3D coordinate at pixel \((u,v)\), $t$ is used to filter low-confidence points. Since the generated pointmaps are per-pixel, we also obtain pixel correspondences between images \(I_{ak} \leftrightarrow I_n\). Propagating from the previously established \cref{eq:corr1}, we construct 2D-3D correspondences between the rendered pointmap of the last keyframe and the current frame image:
\begin{equation}\label{eq:corr2}
    X^r_{ak}\leftrightarrow I_{ak} \leftrightarrow I_{n}
\end{equation}

Using these correspondences, we estimate the relative pose \(\mathbf{T}_n^{\text{rel}}\) between the current frame and the adjacent keyframe via RANSAC \cite{RANSAC} and PnP \cite{lepetit2009epnp}. The key advantage here is that the 3D coordinates in \(X^r_{ak}\) come directly from the 3DGS model, ensuring strict scale consistency with the reconstructed scene. Consequently, the PnP solution inherently preserves the correct scale. 

Finally, the pose of current frame \(n\) is computed as:
\begin{equation}
\mathbf{T}_n = \mathbf{T}_n^{\text{rel}}\,\mathbf{T}_{ak}.
\end{equation}

\subsubsection{Pose Optimization}
\label{Pose Optimization}
To achieve precise camera pose, we leverage the 3DGS differentiable rendering pipeline to generate images and optimize the pose $T_n$ by minimizing the photometric loss:
\begin{equation}
L_{\text{pho}} = \left\lVert I(\mathcal{G}, T) - \bar{I} \right\rVert_{1},
\end{equation}
where $I$ represents a per-pixel differentiable rendering function, generating images through Gaussian $\mathcal{G}$ and camera pose $T$, and $\bar{I}$ is the ground truth image.

To avoid the overhead of automatic differentiation, similar to \cite{yu2025opengs-slam}, we linearize the camera pose $T\in SE(3)$ into its corresponding Lie algebra $\mathfrak{se}(3)$ and explicitly incorporate the gradient of $T$ within the 3DGS CUDA pipeline:
\begin{equation}
\nabla_T L_{\text{pho}} = \frac{\partial L_{\text{pho}}}{\partial r} \cdot \left( \frac{\partial r}{\partial \mu_I} \cdot \frac{\partial \mu_I}{\partial T} + \frac{\partial r}{\partial \Sigma_I} \cdot \frac{\partial \Sigma_I}{\partial T} \right),
\end{equation}
where $r$ denotes the rasterization function, the derivatives of mean $\mu_I$ and covariance $\Sigma_I$ are derived from \cite{matsuki2024gaussian}.

In particular, to enhance accuracy and focus on details in pose optimization, we penalize non-edge and invalid region. 

Based on the Self-Consistent 3DGS Pointmap Tracking, our proposed method enables more precise and robust tracking from 3DGS scene with fewer iterations.

\subsection{Patch-based Pointmap Dynamic Mapping}

In monocular RGB-only SLAM, the lack of geometric information leads to inaccurate scene reconstruction. One solution is to introduce monocular depth priors, but depth estimation suffers from scale drift across frames. This is particularly problematic in unbounded outdoor scenes, where complex environments cause increasing instability in scale. Some works \cite{zhu2024mgs-slam, zhang2024glorie} align depth scales to sparse point clouds from independent tracking modules, but their performance is limited by point cloud quality and they do not form an end-to-end pipeline. Recent work \cite{yu2025opengs-slam} align scale to the initial frame by establishing correspondences between consecutive frames, but this introduces cumulative errors.

We address this issue by using pre-trained pointmap \cite{mast3r} as geometric prior and propose a patch-based method to dynamically align its scale to the current Gaussian scene. Through pointmap replacement, we achieve optimal Gaussian insertion at keyframes. With the aligned pre-trained pointmap, we perform geometric supervision optimization on the Gaussian scene. 

\subsubsection{Patch-Based Scale Alignment}
\label{Patch-Based Scale Alignment}
After optimizing the current frame's pose, we rasterize a 3DGS pointmap $X^r$ from the current viewpoint $T_n$ and obtain the pre-trained pointmap $X^p$. The scale of the $X^r$ is consistent with the scene scale but is usually less precise than the $X^p$. Our approach is to identify reliable pixels in the $X^r$ as ``correct points'' and use them to calculate a scaling factor to align the $X^p$'s scale to the scene.

However, finding ``correct point'' when there is a scale discrepancy is challenging, necessitating preliminary alignment of the two pointmaps. Direct normalization might lead to incorrect identification of ``correct points'' due to inconsistent value distributions or outliers in $X^r$. To address this, we segment the entire pointmap into small patches and select patches with similar distributions for normalization, ensuring accurate point selection from $X^r$. 

We initially segments $X^p$ and $X^r$ into patches of size $P \times P$, then calculates the mean $\mu$ and standard deviation $\sigma$ for each patch. Patches are selected for normalization if they satisfy $
|\mu_r - \mu_p| < \delta_{\mu} \times \mu_p \quad \text{and} \quad |\sigma_r - \sigma_p| < \delta_{\sigma} \times \sigma_p
$.
For these candidate patches, pointmap values are normalized:
\begin{equation}
X_N(x) = \frac{X(x) - \mu(X)}{\sigma(X)},
\end{equation}
Points satisfying 
$
|X^r_N(x) - X^p_N(x)| < \epsilon_r
$
are selected as ``correct points'' set $CP$, and scale factor is calculated:
\begin{equation}
\sigma' = \frac{\mu(X^r[CP])}{\mu(X^p[CP])},
\end{equation}
and applied to adjust $X^p$. This process is iterated until the scale factor stabilizes or the iteration limit is reached, eventually outputting the aligned $\hat{X}^p = \sigma \times X^p$ for pointmap replacement and supervision. If the number of ``correct points'' is insufficient, the scale factor estimation might be erroneous, introducing additional biases in scene reconstruction. In this case, we establish point correspondences with the pre-trained pointmap $\hat{X}^p_{ak}$ from adjacent keyframe and compute a remedial scale factor. Since $\hat{X}^p_{ak}$ is already aligned with the scene, it can serve as a reference. For detailed algorithm and pseudocode, see the supplementary material.

\subsubsection{Pointmap Replacement}
\label{Pointmap Replacement}
We insert new Gaussians into the scene at keyframes. To minimize scale drift, we initialize the Gaussians based on the rendered pointmap $X^r$ from the current frame. However, the rendered pointmap often contains poorly reconstructed areas, and using it directly may introduce additional errors. We use the aligned pre-trained pointmap $\hat{X}^p$ as a reference and replace the ``incorrect points'' in $X^r$:

\begin{footnotesize}
\begin{equation}
\label{point replace}
\hat{X}^r(x)= \begin{cases}
X^r(x), & \text{if } |X^r(x)-\hat{X}^p(x)| \le \epsilon_m \times \hat{X}^p(x) \\
\hat{X}^p(x), & \text{if } |X^r(x)-\hat{X}^p(x)| > \epsilon_m \times \hat{X}^p(x)
\end{cases}.
\end{equation}
\end{footnotesize}
We perform random sparse downsampling on $\hat{X}^r(x)$ to effectively control the number of 3D Gaussians, ensuring high-quality mapping while reducing processing time.

\begin{figure*}
  \centering
  \includegraphics[width=\textwidth]{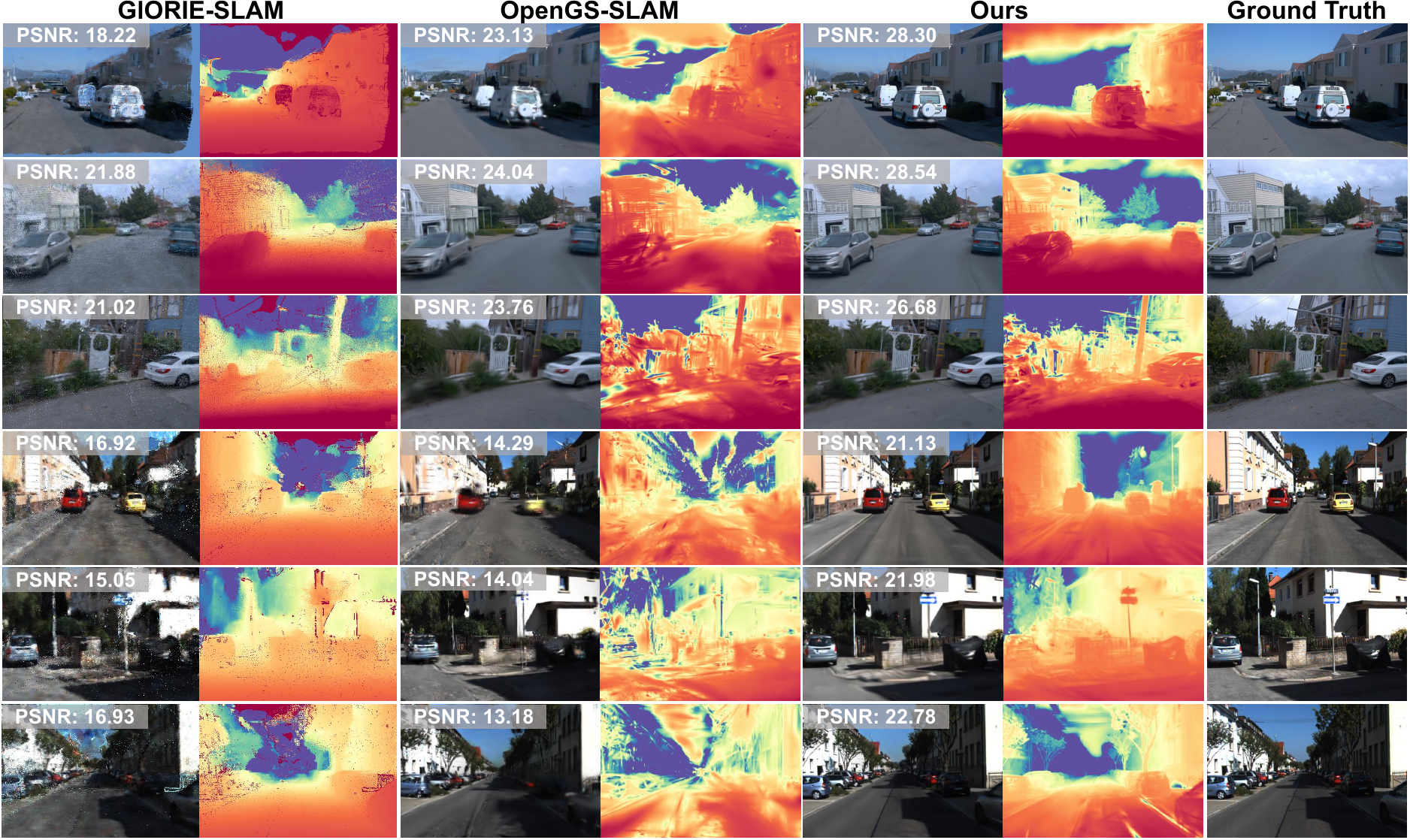}
  \vspace{-15pt}
  \caption{\textbf{Novel View Synthesis Results on Waymo (top three rows) and KITTI (bottom three rows) scenes, including Rendered RGB and Depth Maps.} Our method produces high-fidelity images that capture intricate details of vehicles, streets, and buildings. The rendered depth maps are more accurate in regions with complex depth variations, such as tree branches and roadside vehicles.}
  \label{fig:waymo_render}
  \vspace{-7pt}
\end{figure*}

\begin{table*}[ht]
\centering
\fontsize{8.5}{10}\selectfont  
\setlength{\tabcolsep}{0.7pt} 
\renewcommand{\arraystretch}{1.3} 
    \begin{tabular}{>{\centering\arraybackslash}p{3cm}|>{\centering\arraybackslash}p{1.0cm}*{3}{>{\centering\arraybackslash}p{1.0cm}}|>{\centering\arraybackslash}p{1.0cm}*{3}{>{\centering\arraybackslash}p{1.0cm}}|>{\centering\arraybackslash}p{1.0cm}*{3}{>{\centering\arraybackslash}p{1.0cm}}}
    \toprule
    \multirow{2}{*}{\textbf{Method}} & \multicolumn{4}{c}{\textbf{Waymo} \cite{Sun_2020_CVPRwaymo}} & \multicolumn{4}{c}{\textbf{KITTI} \cite{geiger2013kittidata}} & \multicolumn{4}{c}{\textbf{DL3DV} \cite{ling2024dl3dv}} \\ \cmidrule(lr){2-5} \cmidrule(lr){6-9}  \cmidrule(lr){10-13}
                        & ATE$\downarrow$ & PSNR$\uparrow$ & SSIM$\uparrow$ & LPIPS$\downarrow$ & ATE$\downarrow$ & PSNR$\uparrow$ & SSIM$\uparrow$ & LPIPS$\downarrow$ & ATE$\downarrow$ & PSNR$\uparrow$ & SSIM$\uparrow$ & LPIPS$\downarrow$ \\ \hline
    \textbf{NeRF-SLAM} \cite{rosinol2022nerfslamrealtimedensemonocular}      &  97.36   &  10.12   & 0.597   &  0.781  & 63.55   & 11.33   & 0.441   &  0.774   &  4.41   & 12.21 &  0.505 &  0.517 \\ 
    \textbf{NICER-SLAM}\cite{zhu2024nicer}    & 19.59 & 12.22 & 0.622 & 0.726 & 18.55 & 12.69 & 0.450 & 0.701 & 3.57 & 14.19 & 0.551 & 0.507 \\ 
    \textbf{Photo-SLAM} \cite{huang2024photo}   & 19.95& 17.73 & 0.741 & 0.674 & 17.62 & 15.39 & 0.523 & 0.674 & 2.78 & 16.74 & 0.560 & 0.499 \\ 
    \textbf{GlORIE-SLAM} \cite{zhang2024glorie}   & \textbf{0.589} & 18.83 & 0.702 & 0.572 & 1.134 & 15.49 & 0.594 & 0.684 & 0.492 & 16.20 & 0.669 & 0.515 \\ 
    \textbf{MonoGS} \cite{matsuki2024gaussian}        & 8.529 & 21.80 & 0.780 & 0.577 & 9.493 & 14.78 & 0.486 & 0.759 & 0.274 & 24.99 & 0.766 & 0.322\\ 
    \textbf{OpenGS-SLAM} \cite{yu2025opengs-slam}   & 0.839 & 23.99 & 0.800 & 0.434 & 3.224 & 15.61 & 0.495 & 0.492 & 0.141 & 24.75 & 0.788 & 0.192  \\ 
    \hline
    \textbf{Ours}          & \underline{0.622} & \textbf{26.73} & \textbf{0.845} & \textbf{0.360} & \textbf{1.048} & \textbf{20.03} & \textbf{0.646} & \textbf{0.398} & \textbf{0.032} & \textbf{29.97} & \textbf{0.893} & \textbf{0.108}\\ 
    \bottomrule
    \end{tabular}
    \caption{\textbf{Comparison of all methods on three datasets.} ATE RMSE (m) for tracking, and PSNR, SSIM, LPIPS for Novel View Synthesis. Best results are in \textbf{bold}, second-best in \underline{underlined}. Our method achieves NVS SOTA performance across all datasets, with the best tracking accuracy on KITTI and DL3DV, and comparable tracking accuracy to GlORIE-SLAM on Waymo.}
    \label{table:main_result}
\end{table*}

\subsubsection{Map Optimization with Pointmap Supervision}
\label{Map Optimization with Pointmap Supervision}
To achieve efficient viewpoint coverage and introduce multi-view constraints, inspired by \cite{engel2017dso,matsuki2024gaussian}, we jointly refine camera poses and the Gaussian map within the current local keyframe window $\mathcal{W}$. See the supplementary material for details. 

To improve the scene geometry, we introduce a pointmap-based geometry loss:
\begin{equation}
    L_{geo}= \left\lVert X^r - \hat{X}^p \right\rVert_{1}.
\end{equation}

To mitigate excessive stretching of the ellipsoids and reduce artifacts, we employ isotropic regularization \cite{matsuki2024gaussian}:
\begin{equation}
    L_{\text{iso}} = \sum_{i=1}^{|\mathcal{G}|} \left\lVert s_i - \tilde{s}_i \cdot \mathbf{1} \right\rVert_1,
\end{equation}
where $\tilde{s_i}$ denotes the mean of the scaling $s_i$. Combining the photometric loss, geometry loss, and isotropic regularization, the map optimization task can be summarized as:
\begin{equation}
    \min_{\substack{T_k \in SE(3) \\ \forall k \in \mathcal{W}}, \mathcal{G}} \quad \sum_{\forall k \in \mathcal{W}} \alpha L_{\text{pho}}^k + (1-\alpha)L^k_{geo}+\lambda_{\text{iso}} L_{\text{iso}}.
\end{equation}

\begin{figure*}
  \centering
  \includegraphics[width=1\textwidth]{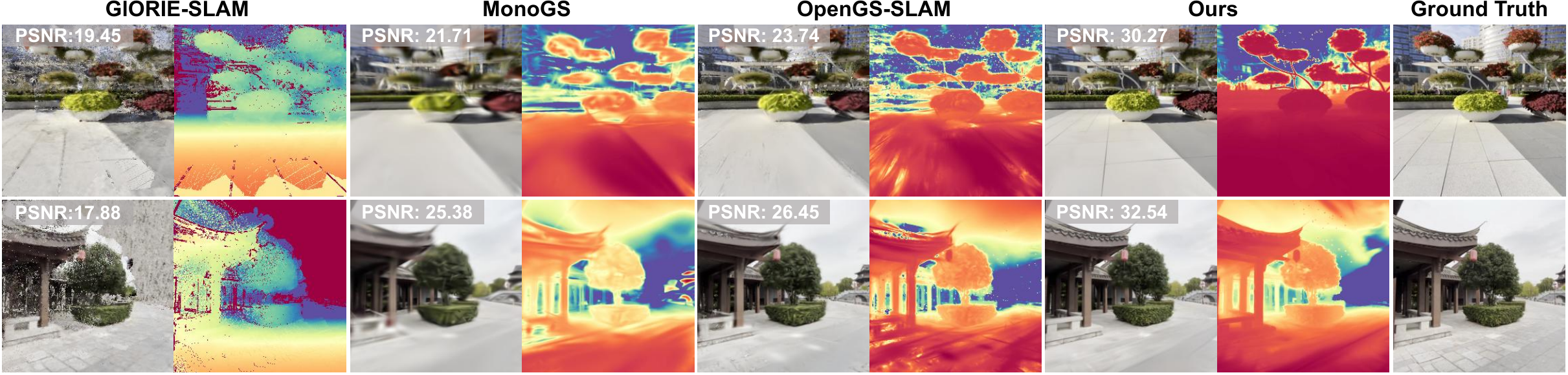}
  \vspace{-5pt}
  \caption{\textbf{Novel View Synthesis Results on DL3DV scenes, including Rendered RGB and Depth Maps.} Our method renders clearer images with better reconstruction of details such as flowerbeds, eaves, and lanterns. The depth maps more accurately capture interwoven objects like columns and show greater stability and smoothness in ground regions.}
  \label{fig:dl3dv_render}
  \vspace{-5pt}
\end{figure*}

\section{Experiments}

\subsection{Implementation and Experiment Setup}

\noindent \textbf{Datasets.} We conduct experiments on the Waymo Open Dataset \cite{Sun_2020_CVPRwaymo}, KITTI Dataset \cite{geiger2013kittidata}, and DL3DV Dataset \cite{ling2024dl3dv} to evaluate tracking accuracy and novel view synthesis performance in outdoor environment. Specifically, we select nine 200-frame sequences from Waymo, eight 200-frame sequences from KITTI, and three 300-frame sequences from DL3DV. The selected scenes are all static and feature significant camera viewpoint changes.

\noindent \textbf{Metrics.} To assess novel view synthesis performance, we use PSNR, SSIM \cite{SSIM}, and LPIPS metrics, calculated on frames excluding keyframes (i.e., training frames). For tracking accuracy, we use ATE RMSE (m) for evaluation.

\noindent \textbf{Baseline Methods.} 
We compare our method with SLAM approaches that support monocular RGB-only input and novel view synthesis. These include NeRF-based methods NeRF-SLAM \cite{rosinol2022nerfslamrealtimedensemonocular} and NICER-SLAM \cite{zhu2024nicer}, implicit encoding point cloud-based GlORIE-SLAM \cite{zhang2024glorie}, 3DGS-based methods MonoGS \cite{matsuki2024gaussian} and Photo-SLAM \cite{huang2024photo}, and OpenGS-SLAM \cite{yu2025opengs-slam}, which is specifically designed for outdoor environments.

\begin{figure}
  \centering
  \includegraphics[width=0.9\columnwidth]{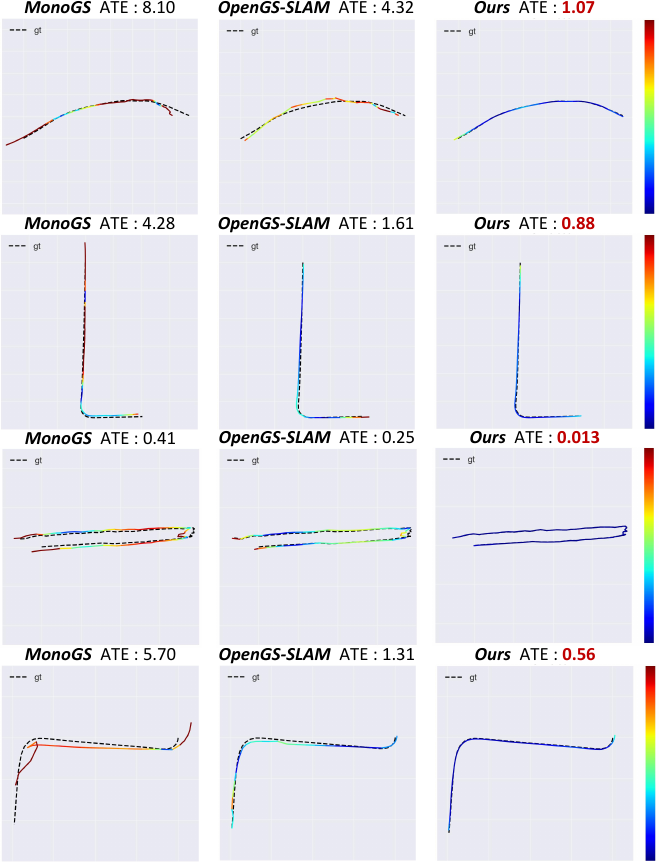}
  \vspace{-5pt}
  \caption{\textbf{Comparison of Tracking Trajectories with OpenGS-SLAM and MonoGS.} Under large viewpoint changes, MonoGS struggles to track, while OpenGS-SLAM exhibits instability. In contrast, our method achieves superior robustness.}
  \label{fig:tracking_traj}
  \vspace{-10pt}
\end{figure}

\noindent \textbf{Implementation Details.} Experiments are conducted on an NVIDIA RTX A6000 GPU. Similar to MonoGS, Gaussian attributes and camera pose rasterization and gradient computations are implemented using CUDA. The remainder of the SLAM pipeline is developed in PyTorch. More details are provided in the supplementary material.

\begin{table}
    \centering
    \fontsize{9}{9}\selectfont  
    \setlength{\tabcolsep}{1pt} 
    \renewcommand{\arraystretch}{1} 
    \newcommand{\greenbox}[1]{\cellcolor{green!20}#1}
    \begin{tabular}{>{\centering\arraybackslash}l *{5}{>{\centering\arraybackslash}p{1.1cm}}}
        \toprule
        \multirow{2}{*}{\textbf{Method}}&  \multicolumn{5}{c}{\textbf{Iterations}} \\
        \cmidrule(lr){2-6}
         & 5 & 15 & 30 & 50 & 100 \\
        \midrule
        MonoGS & 12.6 & 9.87 & 8.95 & 2.98 & 1.70 \\
        OpenGS-SLAM & 4.17 & 2.52 & 1.10 & 0.85 & 0.80 \\
        \midrule
        \textbf{Ours} & \textbf{0.55} & \textbf{0.55} & \textbf{0.51} & \textbf{0.49} & \textbf{0.46} \\
        \bottomrule
    \end{tabular}
    \vspace{-7pt}
    \caption{\textbf{Tracking error (ATE RMSE) on Waymo\_405841 under different iteration counts.} }
    \label{convergence}
    \vspace{-2pt}
\end{table}
\begin{table}
\centering
\fontsize{9}{9}\selectfont  
\setlength{\tabcolsep}{6.8pt} 
\label{table:ablation}
\begin{tabular}{lcccc}
\toprule 
\textbf{Method} & \textbf{ATE} & \textbf{PSNR} & \textbf{SSIM} & \textbf{LPIPS} \\
\midrule 
MonoGS+MASt3R & 3.84 & 23.09 & 0.792 & 0.439 \\
\midrule
\textbf{Ours}  & \textbf{0.62} & \textbf{26.73} & \textbf{0.845} & \textbf{0.360} \\
\bottomrule 
\label{ablation}
\end{tabular}
\vspace{-5pt}
\caption{\textbf{Comparison with direct incorporation of pre-trained pointmap information in 3DGS-based SLAM (e.g., MonoGS).} We report the average results on Waymo.}
\label{table:xr_straight-pointmap}
\vspace{-5pt}
\end{table}

\subsection{Experiment Results}

\noindent \textbf{Camera Tracking.}  
\Cref{table:main_result} presents the tracking results on three datasets. Our method achieves state-of-the-art performance on the KITTI and DL3DV datasets, and comparable performance to GlORIE-SLAM on the Waymo dataset. While GlORIE-SLAM tracks the camera by leveraging the relationship between consecutive image frames, our method, like MonoGS and OpenGS-SLAM, performs visual localization from a single image to the scene for pose estimation, a more challenging task. Despite this, our method achieves comparable tracking accuracy to GlORIE-SLAM on the Waymo and KITTI datasets, and significantly outperforms it on DL3DV.  Compared to OpenGS-SLAM, which is also 3DGS-based and designed specifically for outdoor environments, we reduce tracking error by \textbf{67.5\%} on KITTI and by \textbf{77.3\%} on DL3DV.

\Cref{fig:tracking_traj} shows the comparison of tracking trajectories. In the presence of significant camera displacement and rotation, both MonoGS and OpenGS-SLAM exhibit unstable convergence and poor perception of displacement, while our method demonstrates superior accuracy and robustness. These results highlight the superiority of our method. 

\noindent \textbf{Novel View Synthesis.} 
As shown in \cref{table:main_result}, our method achieves the best novel view synthesis performance across all three datasets. Compared to the current best 3DGS-based SLAM methods, PSNR is significantly improved: \textbf{+2.73} on Waymo, \textbf{+4.42} on KITTI, and \textbf{+4.98} on DL3DV. 

\Cref{fig:waymo_render,fig:dl3dv_render} show the rendered images and depth maps for the three datasets. For outdoor scenes, our method generates high-fidelity images with better reconstruction of vehicle, building, and street details. Additionally, our rendered depth maps are more accurate in regions with complex depth variations and exhibit more reasonable relative positioning between objects. This demonstrates our method’s strong geometric understanding of outdoor scenes and reflects the stability of pointmap scale during training.

\noindent \textbf{Pose Optimization Convergence.} 
\Cref{convergence} shows the impact of different pose optimization iteration counts on tracking error in the Waymo scene. MonoGS fails to converge with fewer than 50 iterations, while OpenGS-SLAM shows a noticeable accuracy drop below 30 iterations. In contrast, our method achieves results comparable to 100 iterations with only 5 iterations, demonstrating the robustness of our self-consistent 3DGS pointmap tracking module.

\subsection{Ablation Study}

\begin{table}
    \centering
    \fontsize{9}{9}\selectfont  
    \setlength{\tabcolsep}{1.5pt} 
    \renewcommand{\arraystretch}{1} 
    \newcommand{\greenbox}[1]{\cellcolor{green!20}#1}
    \begin{tabular}{>{\centering\arraybackslash}l *{5}{>{\centering\arraybackslash}p{1.1cm}}}
        \toprule
        \multirow{2}{*}{\textbf{Method}} &  \multicolumn{5}{c}{\textbf{Iterations}} \\
        \cmidrule(lr){2-6}
        & 5 & 15 & 30 & 50 & 100 \\
        \midrule
        Ours (w/o PAPE) & 16.9 & 8.57 & 5.61 & 4.76 & 3.48 \\
        Ours & \textbf{0.55} & \textbf{0.55} & \textbf{0.51} & \textbf{0.49} & \textbf{0.46} \\
        \bottomrule
    \end{tabular}
    \vspace{-5pt}
    \caption{\textbf{Ablation study of Pointmap Anchored Pose Estimation (PAPE) module on Waymo\_405841.} We present the tracking errors across different iteration counts.}
    \label{table:xr_pose}
    \vspace{-5pt}
\end{table}

\begin{table}
\centering
\fontsize{9}{9}\selectfont  
\setlength{\tabcolsep}{5.5pt} 
\label{table:ablation}
\begin{tabular}{lcccc}
\toprule 
\textbf{Method} & \textbf{ATE} & \textbf{PSNR} & \textbf{SSIM} & \textbf{LPIPS} \\
\midrule 
w/o Pose Optimization & 1.79 & 24.45 & 0.805 & 0.376 \\
w/o Scale Alignment & 3.50 & 23.49 & 0.784 & 0.411 \\
w/o Point Replacement & 1.35 & 25.59 & 0.825 & 0.406 \\
w/o Geometry Loss $\mathcal{L}_{geo}$ & 3.73 & 25.70 & 0.829 & 0.402 \\
\midrule
\textbf{Ours}  & \textbf{0.62} & \textbf{26.73} & \textbf{0.845} & \textbf{0.360} \\
\bottomrule 
\label{ablation}
\end{tabular}
\vspace{-7pt}
\caption{\textbf{Ablation study on key modules.} We report the average results on Waymo.}
\label{table:main_xr}
\vspace{-7pt}
\end{table}

\begin{figure}
  \centering
  \includegraphics[width=0.9\columnwidth]{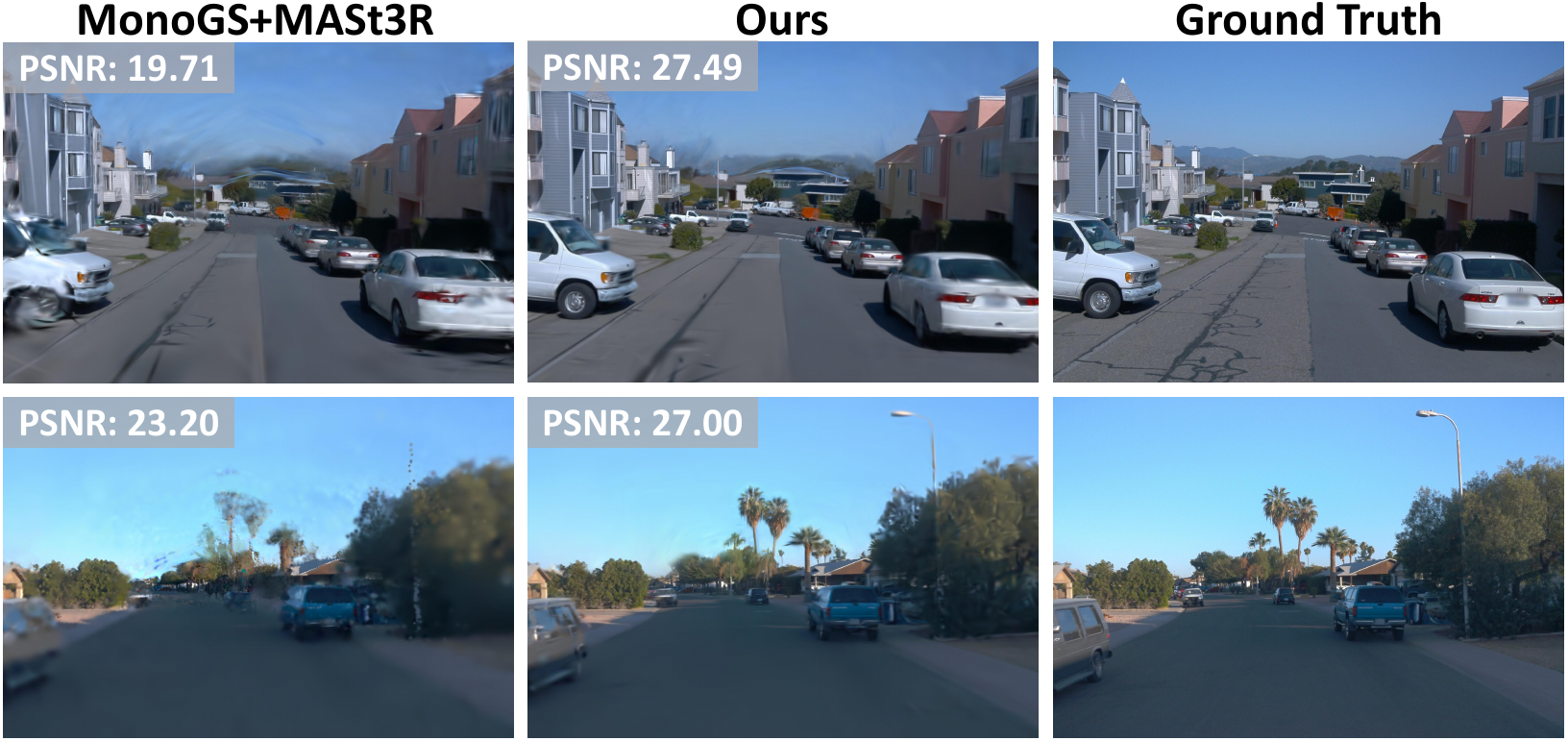}
  \vspace{-5pt}
  \caption{\textbf{Comparison with direct incorporation of pre-trained pointmap information in 3DGS-based SLAM (e.g., MonoGS).} Scale drift leads to noticeable geometric blurring.}
  \label{fig:xr_straight-pointmap}
  \vspace{-10pt}
\end{figure}

\noindent \textbf{Self-Consistent 3DGS Pointmap Tracking.} 
\Cref{table:xr_pose} shows that without the Pointmap Anchored Pose Estimation module, camera tracking fails to converge, especially with fewer optimization iterations, leading to a rapid drop in accuracy. The first row of \cref{table:main_xr} indicates that without pose optimization, both tracking and reconstruction performance degrade. This demonstrates that pose estimation ensures convergence in complex outdoor environments, while pose optimization further refines pose accuracy.

\noindent \textbf{Patch-based Pointmap Dynamic Mapping.} 
As shown in \cref{table:main_xr}, the absence of pointmap replacement leads to erroneous Gaussian insertions in keyframes, degrading both tracking and reconstruction performance. Removing $\mathcal{L}_{geo}$ results in a significant increase in tracking error due to the lack of geometric supervision, while reconstruction quality only slightly declines. This is because pointmap replacement provides geometric priors, allowing reasonable reconstruction quality even when relying solely on photometric loss. However, this compromises the system’s displacement awareness. Removing patch-based scale alignment causes a substantial performance drop, as misaligned pre-trained pointmaps introduce incorrect supervision.

\noindent \textbf{Pre-trained Pointmap Processing.} 
\Cref{table:xr_straight-pointmap} shows that directly incorporating pre-trained pointmap supervision into 3DGS SLAM (e.g., MonoGS) without our proposed pointmap processing modules results in significantly degraded performance. As seen in \cref{fig:xr_straight-pointmap}, scale drift causes substantial blurring. This demonstrates that pre-trained geometric priors alone are insufficient for handling outdoor scenes, and the proposed designs are the key to our method’s success.

\section{Conclusion}

In this work, we introduce S3PO-GS, a monocular outdoor 3D Gaussian Splatting SLAM framework with a scale self-consistent pointmap, addressing the challenges of scale drift and lack of geometric priors in outdoor scenes. By using a self-consistent 3DGS pointmap tracking module, we reduce pose estimation iterations to 10\% of traditional methods, achieving centimeter-level tracking accuracy on complex datasets like Waymo. A patch-based dynamic mapping mechanism, based on local patch matching, resolves monocular depth scale ambiguity and enhances reconstruction quality. Experiments show our method sets new benchmarks in tracking accuracy and novel view synthesis for 3DGS SLAM. Future work will explore loop closure and large-scale scene optimization to expand its application boundaries in outdoor SLAM.

\clearpage
\section*{Acknowledgment}
This research is supported by the National Natural Science Foundation of China (No. 62406267), Guangzhou-HKUST(GZ) Joint Funding Program (Grant No.2025A03J3956 \& Grant No.2023A03J0008), the Guangzhou Municipal Science and Technology Project (No. 2025A04J4070), and the Guangzhou Municipal Education Project (No. 2024312122).

{
    \small
    \bibliographystyle{ieeenat_fullname}
    \bibliography{main}
}

\clearpage
\setcounter{page}{1}
\maketitlesupplementary

\section{Overview}
This supplementary material provides implementation details for keyframe management, patch-based scale alignment, pointmap replacement, and Gaussian map optimization modules. We also include additional experiments on KITTI, with runtime, memory, and patch size analysis. Furthermore, we present extra qualitative results on three datasets, including tracking trajectory and novel view synthesis. Finally, we discuss limitations and future work.

\section{Implementation Details}

\subsection{Keyframe Management}
As described in \Cref{Map Optimization with Pointmap Supervision}, we joinly refine camera poses and the Gaussian map within a local keyframe window $\mathcal{W}$. A well-designed keyframe selection strategy must ensure sufficient viewpoint coverage while avoiding redundancy. Given the computational cost of jointly optimizing the Gaussian scene and camera pose across all keyframes, we maintain a local keyframe window $\mathcal{W}$ to select nonredundant keyframes that observe overlapping areas of the scene. This approach provides better multi-view constraints for subsequent Gaussian map optimization. With this in mind, we adopt the keyframe management approach from \cite{matsuki2024gaussian}, where keyframes are selected based on covisibility, and the local window is managed by assessing the overlap with the latest keyframe.

Specifically, we define the covisibility and overlap between keyframes $i$ and $j$ using Intersection over Union (IOU) and Overlap Coefficient (OC) \cite{matsuki2024gaussian}:

\begin{equation}
IOU_{\text{cov}}(i, j) = \frac{\left| \mathcal{G}_i^v \cap \mathcal{G}_j^v \right|}{\left| \mathcal{G}_i^v \cup \mathcal{G}_j^v \right|},
\end{equation}

\begin{equation}
\quad OC_{\text{cov}}(i, j) = \frac{\left| \mathcal{G}_i^v \cap \mathcal{G}_j^v \right|}{\min\left( \left| \mathcal{G}_i^v \right|, \left| \mathcal{G}_j^v \right| \right)},
\end{equation}
where $\mathcal{G}_i^v$ is the set of visible Gaussians in keyframe $i$. 

Given the latest keyframe $j$, keyframe $i$ is added to the keyframe window $\mathcal{W}$ if: $IOU_{\text{cov}}(i, j)<k_{I}$ or the relative pose translation distance $d_{ij}>k_d\hat{D}_i$, where $\hat{D}_i$ represents the median pointmap depth of frame $i$. Given the newly added keyframe $i'$, we remove keyframe $l$ from the window if: $OC{\text{cov}}(i', l)<k{o}$. If the number of keyframes in the window $\mathcal{W}$ exceeds the maximum size, we remove the keyframe with the lowest $OC$ value relative to $i'$.

For all experiments on three datasets, we set the keyframe management parameters as $k_I=0.9, k_d=0.08, k_o=0.3$, with the keyframe window size set to $|\mathcal{W}|=8$.

\subsection{Patch-based Scale Alignment}

As described in Section \Cref{Patch-Based Scale Alignment}, we align the scale of the pretrained pointmap $X^p$ to Gaussian scene, using the 3DGS pointmap $X^r$ as reference. We propose a rigorous and detailed patch-based method to select highly reliable ``correct points'' and use them to calculate the scale factor. The detailed procedure is described in the following \cref{sup:psu}:

\vspace{-3pt}
\begin{algorithm}
\caption{Patch-based Pointmap Scale Alignment}
\label{sup:psu}
\begin{algorithmic}[1] 
\Procedure{Align}{$X^r, X^p, P, \delta_\mu, \delta_\sigma, \epsilon_r, \text{max\_iter}$}
    \State $\sigma' \gets 1$
    \State $X^p_1 \gets X^p$
    \For{$\text{iter} = 1$ \textbf{to} $\text{max\_iter}$}
        \State \textit{Segment $X^r$ and $X^p_{iter}$ into $P \times P$ patches}
        \For{\textbf{each} patch in $X^r, X^p_{iter}$}
            \State $\mu_r, \sigma_r \gets \text{mean}(X^r), \text{std}(X^r)$
            \State $\mu_p, \sigma_p \gets \text{mean}(X^p_{iter}), \text{std}(X^p_{iter})$
            \If{$|\mu_r - \mu_p| < \delta_\mu \cdot \mu_p \land |\sigma_r - \sigma_p| < \delta_\sigma \cdot \sigma_p$}
                \State \textit{Add patch to candidates}
            \EndIf
        \EndFor
        \For{\textbf{each} patch in candidates}
            \State $X^r_N, X^p_N \gets \frac{X^r - \mu_r}{\sigma_r}, \frac{X^p - \mu_p}{\sigma_p}$
            \For{\textbf{each} $x$ in patch}
                \If{$|X^r_N(x) - X^p_N(x)| < \epsilon_r$}
                    \State \textit{Add $x$ to $CP$}
                \EndIf
            \EndFor
            \If{$CP$ \textbf{is not} empty}
                \State $\sigma' \gets \frac{\mu(X^r[CP])}{\mu(X^p[CP])}$
            \EndIf
        \EndFor
        \State $X^p_{\text{iter}+1} \gets \sigma' \cdot X^p$
    \EndFor
    \State \Return $\hat{X}^p = \sigma' \cdot X^p$
\EndProcedure
\end{algorithmic}
\end{algorithm}
\vspace{-3pt}

If the number of ``correct points'' is insufficient, i.e., $|CP| < \tau N_p$, where $N_p$ represents the number of points in the pointmap, we apply a scale remedy strategy. Specifically, we use the fast NN algorithm \cite{mast3r} to establish matching points $MP$ between the current frame $X^p_n$ and the adjacent keyframe aligned pointmap $\hat{X}^p_{ak}$. Since $\hat{X}^p_{ak}$ is already aligned with the scene scale, it serves as a reference to calculate the scale factor for $X^p_n$:
\begin{equation}
    \sigma' \gets \frac{\mu(\hat{X}^p_{ak}[MP])}{\mu(X^p_n[MP])}.
\end{equation}
We believe that our carefully designed \cref{sup:psu} provides the most reliable scale factor. Therefore, we first compute $X^p_{\text{max\_iter}+1} \gets \sigma' \cdot X^p$ and then perform an additional iteration of the Patch-based Scale Alignment process to obtain a newly estimated scale factor $\sigma''$. If the number of ``correct points'' is sufficient, we adopt this iteration’s result as the final output for scale alignment. However, if the number remains insufficient, we use $\sigma''$ as a remedial scale factor. Although this is not the ideal scenario, it still provides an adequate scale correction for the pre-trained pointmap $X^p$, effectively mitigating severe scale drift. Moreover, experiments show that the number of keyframes requiring a remedial scale factor does not exceed three per scene on average.

The parameter selection for the algorithm is shown in Table \ref{para_align}. Across three outdoor datasets, our method operates with the same parameters without requiring additional tuning for different scenes, demonstrating its robustness and generalizability.

\begin{table}
\centering
\fontsize{9}{9}\selectfont  
\setlength{\tabcolsep}{6pt} 
\label{table:ablation}
\begin{tabular}{lcccccc}
\toprule 
\textbf{Parameter} & $P$ & $\delta_\mu$ & $\delta_\sigma$ & $\epsilon_r$ & max\_iter & $\tau$ \\
\midrule 
\textbf{Value} & 10 & 0.3 & 0.3 & 0.1 & 3 & 0.01 \\
\bottomrule 
\vspace{-10pt}
\end{tabular}
\caption{\textbf{Hyperparameters for Patch-Based Scale Alignment on three datasets.}}
\vspace{-10pt}
\label{para_align}
\end{table}

\subsection{Pointmap Supervision}
To avoid inserting Gaussians at incorrect positions at keyframes, we replace ``incorrect points'' in the rendered pointmap $X^r$ with the aligned pretrained pointmap $\hat{X}^p$, as shown in Section \ref{Pointmap Replacement}. For all three datasets, we set $\epsilon_m=0.15$ to replace points with significant discrepancies.

Notably, when camera viewpoint changes are relatively mild and the Gaussian scene within view remains largely complete (e.g., in straight-line trajectories), the proportion of replaced points is around 10\%, ensuring consistent scale for newly inserted Gaussians. In contrast, when viewpoint changes are large and the Gaussian scene has deficiencies (e.g., during sharp turns), the replacement ratio increases to 30\%-50\%. In such cases, the priority is to prevent inserting outlier Gaussians, while the aligned pre-trained pointmap is sufficient to maintain scale consistency. This demonstrates the dynamic adaptability of our method to complex environments. Additionally, we incorporate the Gaussian pruning approach proposed by \cite{matsuki2024gaussian} to remove outlier Gaussians during map optimization.

\subsection{Gaussian Map Optimization}
In Section \ref{Map Optimization with Pointmap Supervision}, we optimize the Gaussian map within the keyframe window $\mathcal{W}$. For three datasets, we set $\lambda_{iso}=10, \alpha=0.98$. In relatively confined scenes where pointmap values exhibit limited variation, we recommend using a smaller $\alpha$, such as 0.96.

\begin{table}
    \centering
    \fontsize{9}{9.5}\selectfont  
    \begin{tabular}{l|c|c|c|c}
    \toprule
     Method & ATE & PSNR & GPU & Time \\
     \midrule
     DROID-SLAM   & 1.30 & - & 11 G  & $\sim$ 1.5 min\\
     MASt3R-SLAM   & 1.35 & - & 7 G & $\sim$2 min \\
     MonoGS  &  5.68 & 13.3 & 9 G & $\sim$5 min \\
     OpenGS-SLAM  & 1.41 & 17.9 & 9 G & $\sim$10 min \\
     CF-3DGS   & 5.99 & 15.9 & 12 G & $\sim$ 120 min \\
     Splat-SLAM   & 1.25 & 19.5 & 10.5 G & $\sim$36 min \\
    \midrule
    \textbf{Ours} & \textbf{0.55} & \textbf{20.6} & \textbf{9.5 G}  & \textbf{$\sim$5 min}   \\
    \bottomrule
    \end{tabular}
    \vspace{-5pt}
    \caption{Added Comparison on KITTI.}
    \vspace{-10pt}
    \label{tab1}
\end{table}

\begin{table}
    \centering
    \fontsize{9}{9.5}\selectfont  
    \setlength{\tabcolsep}{3pt} 
    \begin{tabular}{l|c|c|c|c|c|c|c|c}
    \toprule
     Patch Size & 5 & 8 & 10 & 12 & 16 & 20 &25 &30\\
    \midrule
     ATE   &  1.35 & 0.62 & 0.55 & 0.57 & 0.49 & 0.69 & 0.73 & 0.97\\
     PSNR   &  18.9 & 20.1 & 20.6 & 20.5 & 20.8 & 20.0 & 20.1 & 19.5\\
    \bottomrule
    \end{tabular}
    \vspace{-5pt}
    \caption{Impact of patch size on KITTI.}
    \label{tab2}
    \vspace{-10pt}
\end{table}

\section{Additional Experiments}
We conducted additional experiments on the KITTI-07 sequence, including further comparisons with CF-3DGS \cite{fu2024cf-3dgs} , MASt3R-SLAM \cite{mast3r}, DROID-SLAM \cite{murai2025mast3r-slam}, and Splat-SLAM \cite{sandstrom2024splat-slam}. We also analyzed runtime and memory consumption, and performed an ablation study on the patch size used in \cref{sup:psu}.
\subsection{Added Comparison}
Tab.~\ref{tab2} shows that CF-3DGS performs significantly worse on the KITTI dataset, while our method achieves notably higher tracking accuracy compared to both DROID-SLAM and MASt3R-SLAM. These improvements stem from our targeted design tailored to the characteristics of outdoor environments and a specific remedy for the scale issues in the MASt3R framework.

\subsection{Running Time and Memory}
Tab.~\ref{tab2} shows that, compared to other 3DGS-based SLAM methods, our approach achieves higher accuracy while maintaining acceptable runtime and memory consumption. Although Splat-SLAM also achieves competitive novel view synthesis (NVS) accuracy, its extensive global optimization procedures incur significant additional runtime.
Note: To ensure high-quality rendering and fair comparison, all the above 3DGS-based methods perform approximately 10 minutes of color refinement after SLAM execution. The reported runtime includes only the full SLAM pipeline, excluding post-processing.

\subsection{Ablation Study on Patch Size}
Tab.~\ref{tab2} shows that a patch size of 10–16 is optimal: larger patches introduce too many outliers, while smaller ones yield noisy statistics.

\section{Additional Qualitative Results}
\begin{figure}
  \centering
  \includegraphics[width=0.9\columnwidth]{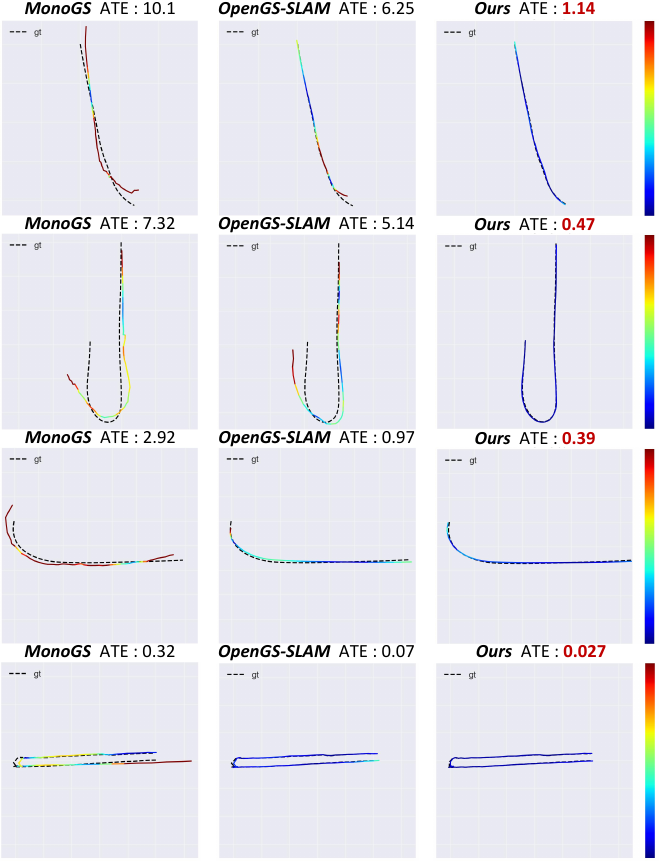}
  \caption{\textbf{Comparison of Tracking Trajectories with MonoGS and OpenGS-SLAM.}}
  \label{fig:tracking_supp}
  \vspace{-5pt}
\end{figure}

\Cref{fig:tracking_supp} presents additional trajectory comparisons, further highlighting the robustness of our method in location under challenging outdoor environments.

\begin{figure*}
  \centering
  \includegraphics[width=\textwidth]{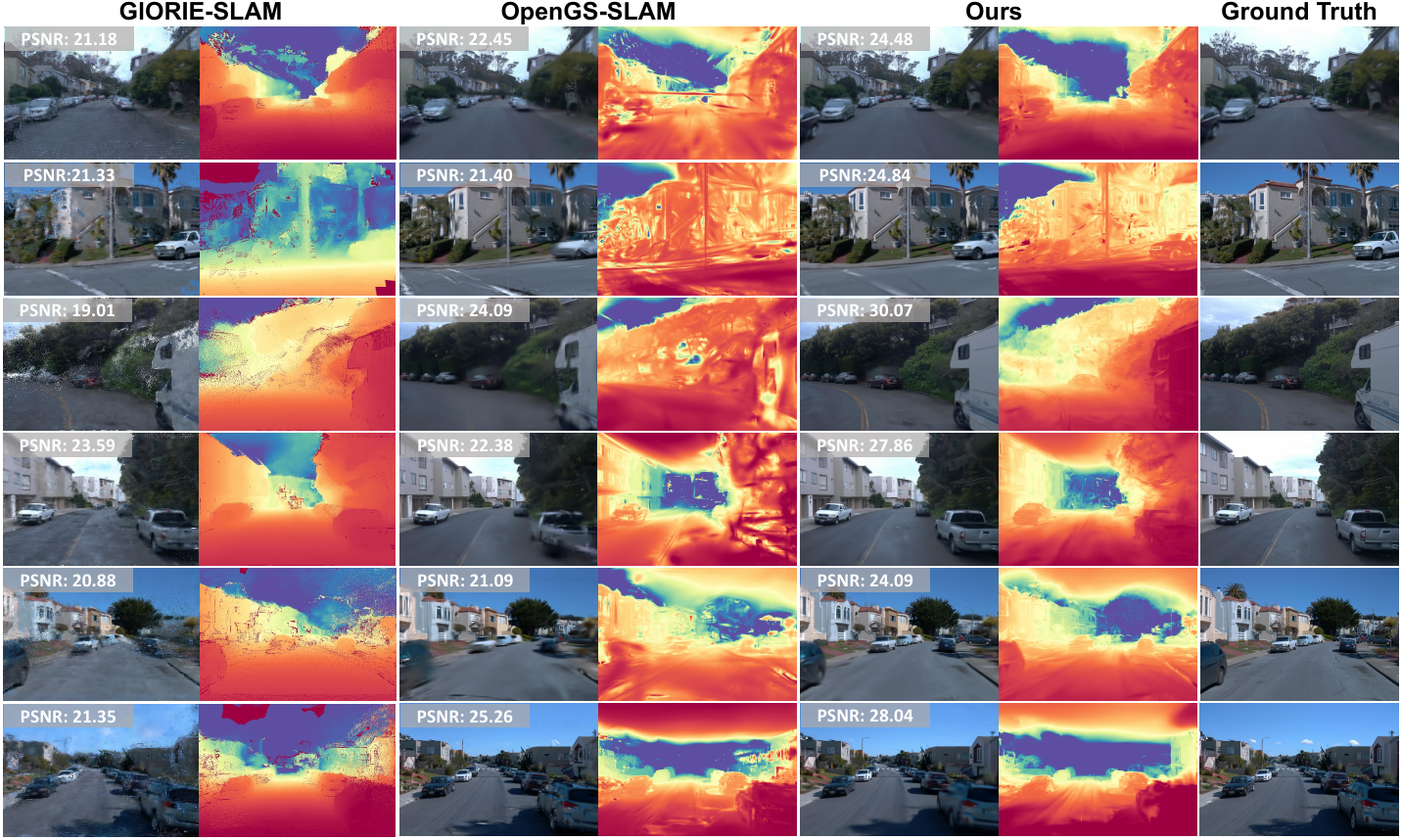}
  \vspace{-10pt}
  \caption{\textbf{Novel View Synthesis Results on Waymo, including Rendered RGB and Depth Maps.}}
  \label{fig:waymo_render_supp}
\end{figure*}

\begin{figure*}
  \centering
  \includegraphics[width=\textwidth]{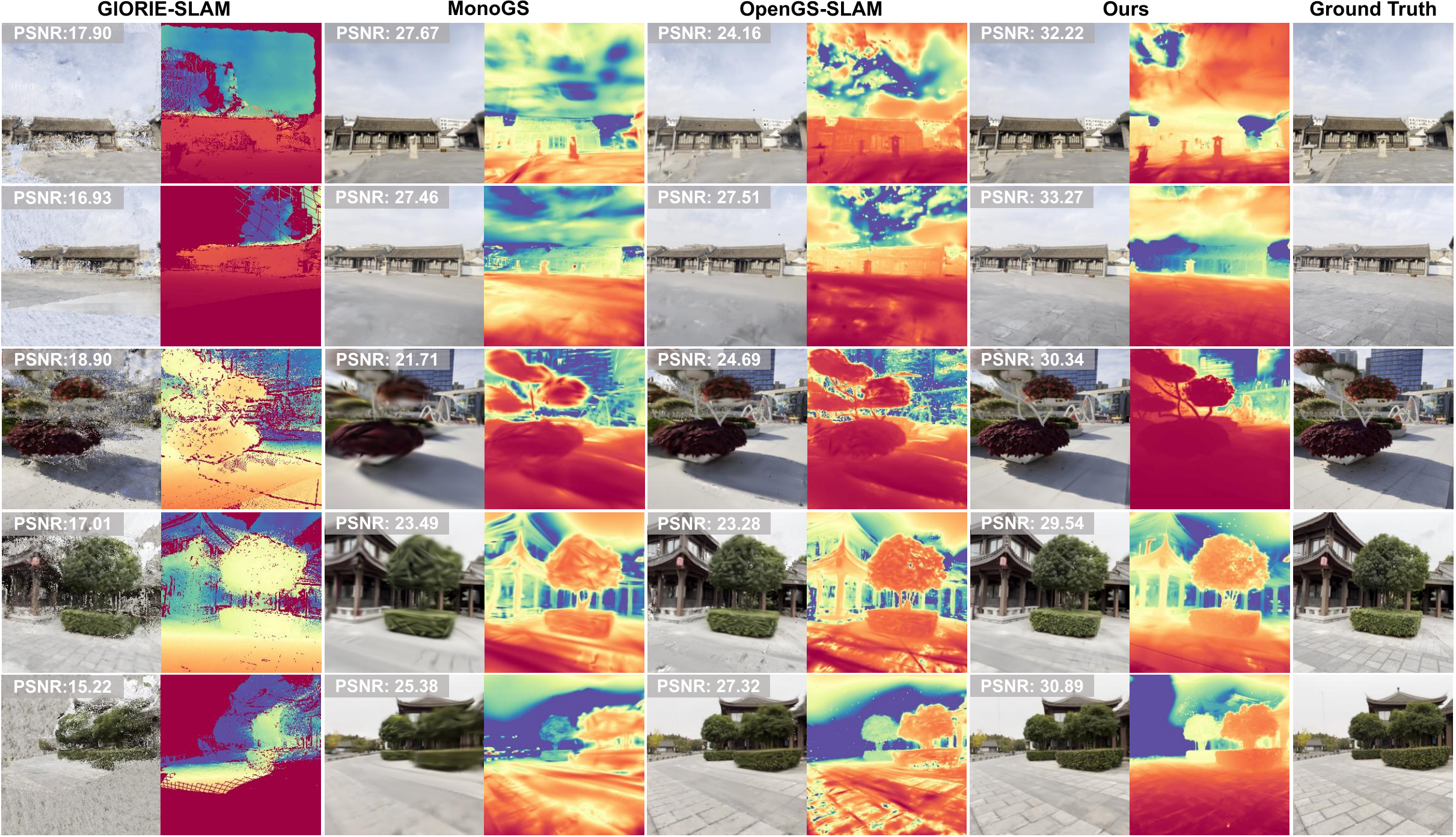}
  \vspace{-10pt}
  \caption{\textbf{Novel View Synthesis Results on DL3DV, including rendered RGB and depth maps.}}
  \label{fig:dl3dv_render_supp}
  \vspace{-10pt}
\end{figure*}

\begin{figure*}
  \centering
  \includegraphics[width=\textwidth]{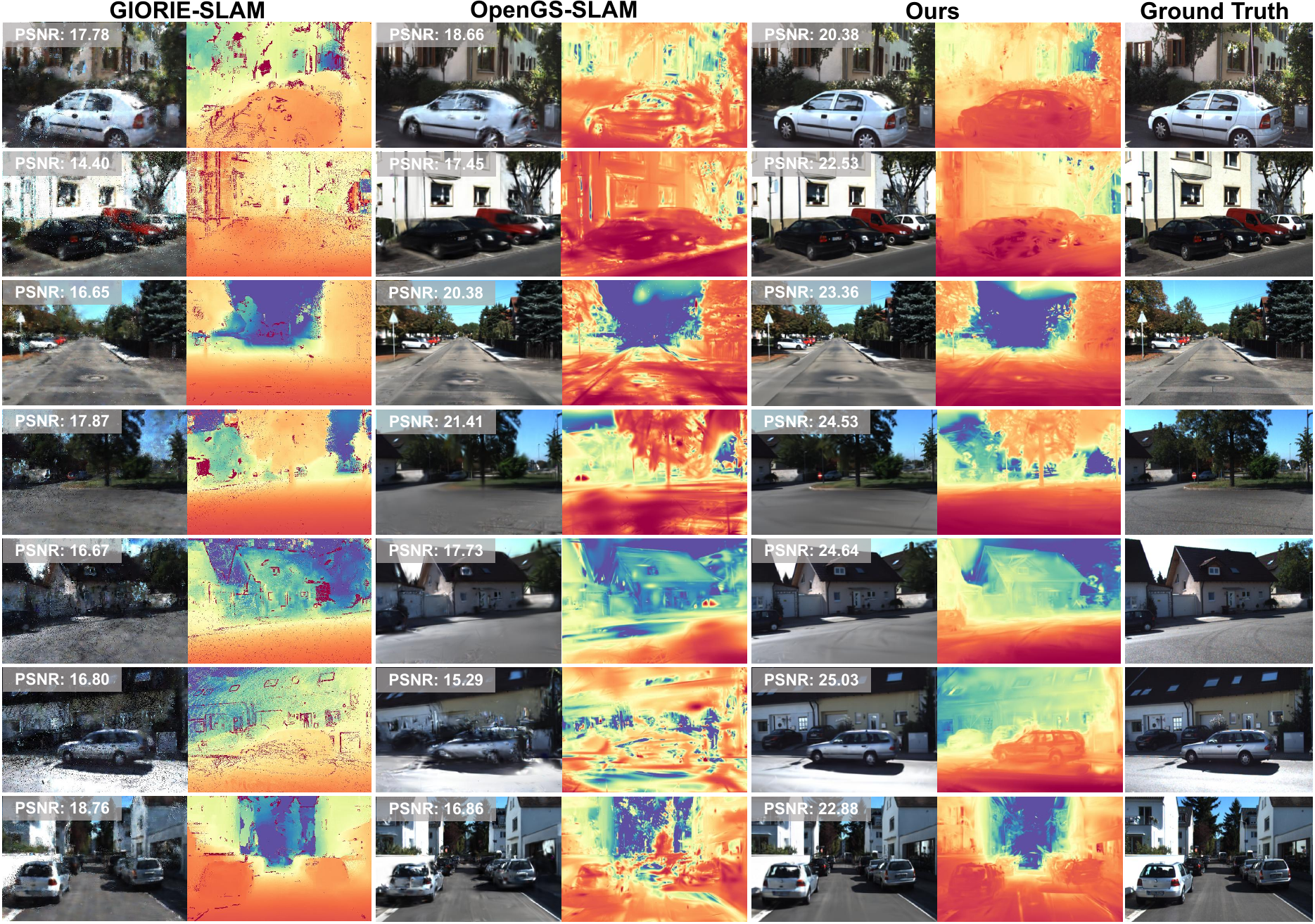}
  \vspace{-10pt}
  \caption{\textbf{Novel View Synthesis Results on KITTI, including Rendered RGB and Depth Maps.}}
  \label{fig:kitti_render_supp}
\end{figure*}

\Cref{fig:waymo_render_supp,fig:kitti_render_supp,fig:dl3dv_render_supp} shows additional novel view synthesis results in the Waymo, DL3DV, and KITTI datasets. Clearly, our method produces higher-fidelity images and more accurate depth maps.

\section{Limitations and Future Works}
\begin{enumerate}
    \item Our method cannot handle dynamic objects in outdoor scenes. Monocular RGB-only SLAM for outdoor environments with dynamic objects remains a highly interesting and challenging problem.
    \item Our method does not incorporate loop closure or global BA. While their inclusion would benefit long-sequence SLAM, it also introduces challenges related to training time and memory consumption.
\end{enumerate}

\end{document}